\documentclass[twoside]{article} 

\usepackage[accepted]{aistats2017}
\usepackage{xcolor}
\definecolor{mydarkblue}{rgb}{0,0.15,0.65}
\usepackage[colorlinks, urlcolor=mydarkblue, citecolor=mydarkblue, linkcolor=mydarkblue]{hyperref}

\usepackage{graphicx} 
\usepackage{subfigure} 

\usepackage{natbib}

\usepackage{algorithm}
\usepackage{algorithmic}

\usepackage{graphicx} 

\usepackage{amsmath}
\usepackage{amssymb}
\usepackage{amsfonts,amsthm}

\usepackage{environ}
\NewEnviron{myequation}{%
\begin{equation}
\scalebox{0.999}{$\BODY$}
\end{equation}
}

\usepackage{xcolor}

\newcommand{\E}{\mathbb{E}}
\newcommand{\R}{\mathbb{R}}
\DeclareMathOperator*{\diag}{diag}
\DeclareMathOperator*{\argmax}{\arg\!\max}

\newcommand{\vv}[1]{\boldsymbol{#1}}

\begin{document}

%

%

\twocolumn[

\aistatstitle{Belief Propagation in Conditional RBMs for Structured Prediction}

\aistatsauthor{ Wei Ping \And Alexander Ihler  }
\aistatsaddress{ Computer Science, UC Irvine \And Computer Science, UC Irvine} 
\vspace{-2.2em}
\aistatsaddress{ weiping.thu@gmail.com \And ihler@ics.uci.edu} 
]

\begin{abstract}
Restricted Boltzmann machines~(RBMs) and conditional RBMs~(CRBMs) are popular models for a wide range of applications.
In previous work, learning on such models has been dominated by contrastive divergence~(CD) and its variants.
Belief propagation~(BP) algorithms are believed to be slow for structured prediction on conditional RBMs~(e.g., \cite{mnih11}), and not as good as CD when applied in learning~(e.g., \cite{larochelle12}). 
In this work, we present a matrix-based implementation of belief propagation algorithms on CRBMs, 
which is easily scalable to tens of thousands of visible and hidden units.
%
We demonstrate that, in both maximum likelihood and max-margin learning, training conditional RBMs with BP as the inference routine can provide significantly better results than current state-of-the-art CD methods on structured prediction problems.
We also include practical guidelines on training CRBMs with BP, and some insights on the interaction of learning and inference algorithms for CRBMs.
\end{abstract}

\section{INTRODUCTION}
\label{sec:introduction}
A restricted Boltzmann machine (RBM) is a two-layer latent variable model that uses a layer of hidden units $\vv h$ to model the distribution of visible units $\vv v$. 
RBMs are widely used as building blocks for deep generative models, such as deep belief networks \citep{hinton06} and deep Boltzmann machines \citep{salakhutdinov09}.
Due to the intractability of the {partition function} in maximum likelihood estimation~(MLE), RBMs are usually learned using the contrastive divergence~(CD) algorithm~\citep{hinton2002CD}, which approximates the gradient of the log-partition function using a k-step Gibbs sampler~(referred to as CD-k). 
To speed up the convergence of the Markov chain, a critical trick in CD-k is to initialize the state of the Markov chain with each training instance.
Although it has been shown that CD-k does not follow the gradient of any objective function~\citep{sutskever2010CD}, 
it works well in many practical applications~\citep{hinton2010practical}.
An important variant of CD-k is persistent~CD~(PCD)~\citep{tieleman2008PCD}.
PCD uses a persistent Markov chain during learning, where the Markov Chain is not reset between parameter updates.
Because the learning rate is usually small and the model changes only slightly between parameter updates, the long-run persistent chain in PCD usually provides a better approximation to the target distribution than the limited step chain in CD-k.

A conditional RBM~(CRBM) is the discriminative extension of RBM to include observed features $\vv x$; 
CRBM is used 
in deep probabilistic model for supervised learning~\citep{hinton06}, 
and also provides a stand-alone solution to a wide range of problems such as classification \citep{larochelle08}, human motion capture~\citep{taylor06}, collaborative filtering~\citep{salakhutdinov07},
and structured prediction~\citep{mnih11,yang14}. 
For structured prediction, a CRBM need not make any explicit assumptions about the structure of the output 
variables~(visible units $\vv v$). 
This is especially useful in many applications where the structure of the outputs is challenging to describe~(e.g., multi-label learning \citep{Li15multilabel}).
In image denoising or image segmentation, the hidden units can encode higher-order correlations of visible units (e.g. shapes, or parts of object), which play the same role as high-order potentials but can improve the statistical efficiency.

In contrast to the success of CD methods for RBMs, it has been noted that both CD-k and PCD may not be well suited to learning conditional RBMs \citep{mnih11}.
In particular, 
PCD is not appropriate for learning such conditional models, because the observed features $\vv x$ greatly affect the model potentials.
 This means we need to run a separate persistent chain for every training instance, which is costly for large datasets.
To make things worse, as we revisit a training instance in stochastic gradient descent~(SGD) (which is standard practice for large datasets), the model parameters will have changed substantially, making the persistent chain for this instance far from the target distribution.
Also, given the observed features, CRBMs tend to be more peaked than RBMs in a purely generative setting.
CD methods may make slow progress because it is difficult for the sampling procedure to explore these peaked but multi-modal distributions.
It was also observed that the important trick in CD-k, which initializes the Markov chain using the training data, does not work well for CRBMs in structured prediction~\citep{mnih11}. In contrast, starting the Gibbs chain with a random state (which resembles the original learning algorithm for Boltzmann machines~\citep{ackley1985learning}) provides better results.

Approximate inference methods, such as mean field~(MF) and belief propagation~(BP), 
can be employed as inference routines in learning
as well as for making predictions after the CRBM has been learned ~\citep{welling2003approximate, yasuda2009approximate}. 
Although loopy BP usually provides a better approximation of marginals than MF~\citep{murphy99loopyBP}, 
it was found to be slow on CRBMs for structured prediction and only considered practical on problems with 
few visible and hidden nodes~\citep{mnih11,mandel2011}.
This inefficiency prevents it from being widely applied to conditional RBMs for structured prediction, in which the CRBMs may have thousands of visible and hidden units.
More importantly, there is a pervasive opinion that belief propagation does not work well on RBM-­based models, 
especially for learning~\citep[][Chapter 16]{Goodfellow-et-al-2016-Book}.

In this work, we present an efficient implementation of belief propagation algorithms for conditional RBMs.
It takes advantage of the bipartite graph structure and is scalable to tens of thousands of visible units and hidden units.
\footnote{For random RBMs with $10,000$ visible units and $2,000$ hidden units, our Matlab implementation converges within a few seconds on a desktop with Intel Core i7~(3.6 GHz).}
Our algorithm uses a compact representation and only depends on matrix product and element-wise operations, which are typically highly optimized in modern high-performance computing architectures.
We demonstrate that, in the conditional setting, learning RBM-based models with belief propagation and its variants can provide much better results than the state-of-the-art CD methods.
We also show that the marginal structured SVM~(MSSVM;~\citep{ping14}) 
can provide improvements for 
max-margin learning of CRBMs~\citep{yang14}.
We include practical guidelines on training CRBMs, and some insights on the interaction of learning and message-passing algorithms for CRBMs.

We organize the rest of the paper as follows.
Section~\ref{sec:relatedwork} discusses some connections to related work.
We review the RBM model and conditional RBMs in Section~\ref{sec:model} and discuss the learning algorithms in Section~\ref{sec:learning}.
In Section~\ref{sec:inference}, we provide our efficient inference procedure. 
We report experimental results in Section~\ref{sec:experiment} and conclude the paper in Section~\ref{sec:conclusion}.

\vspace{-0.4em}
\section{RELATED WORK}
\vspace{-0.5em}
\label{sec:relatedwork}
\citet{mnih11} proposed the CD-PercLoss algorithm for conditional RBMs, 
which uses a CD-like stochastic search procedure to minimize the perceptron loss on training data. 
Given the observed features of the training 
instance, CD-PercLoss starts the Gibbs chain using the logistic regression 
component of the CRBM.
\citet{yang14} trained CRBMs using a latent structured SVM~(LSSVM) objective~\citep{yu2009learning}, and used a greedy search (i.e., iterated conditional modes) for joint maximum a posteriori~(MAP) inference over 
hidden and visible units.

It is also feasible to apply the mean-field~(MF) approximation for the {partition function} in MLE learning of RBMs and CRBMs~\citep{peterson1987}.
Although efficient, this is conceptually
 problematic in the sense that it effectively maximizes an upper bound of the log-likelihood in learning.
In addition, MF uses a unimodal proposal to approximate the multi-modal distribution, which may lead to unsatisfactory results.

Although belief propagation~(BP) and its variants have long been used to learn conditional random fields~(CRFs) with hidden variables~\citep{quattoni2007hidden,ping14}, 
they are mainly applied on sparsely connected graphs (e.g., chains and grids) and were believed to be ineffective and slow on very dense graphs like CRBMs~\citep{mnih11, Goodfellow-et-al-2016-Book}.
A few recent works 
impose particular assumptions on the type of edge potentials and provide efficient inference algorithms for fully connected CRFs. 
For example, the edge potentials in \citep{krahenbuhl2012efficient} are defined by a linear combination of Gaussian kernels.
In this work, however, we propose to speed up general belief propagation on conditional RBMs without any potential function restrictions.

\vspace{-0.4em}
\section{MODELS}
\vspace{-0.5em}
\label{sec:model}
In this section, we review background on RBMs and conditional RBMs.
We also discuss structured prediction with CRBMs.
\subsection{Restricted Boltzmann Machine}
\begin{figure}[t]
\centering
\begin{tabular}{cc}
\!\!{\includegraphics[height=.12\textwidth]{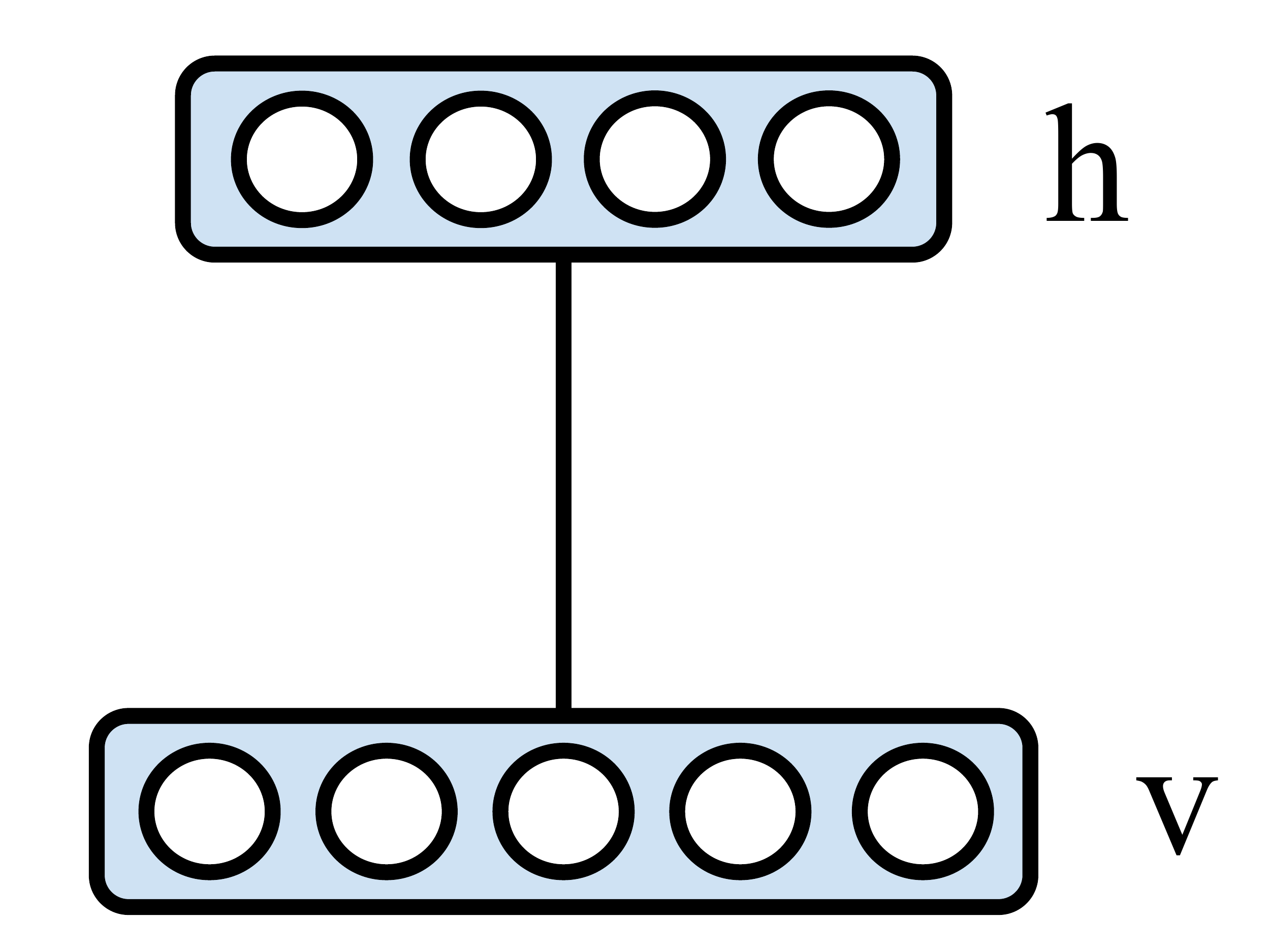}}
& 
{\includegraphics[height=.12\textwidth]{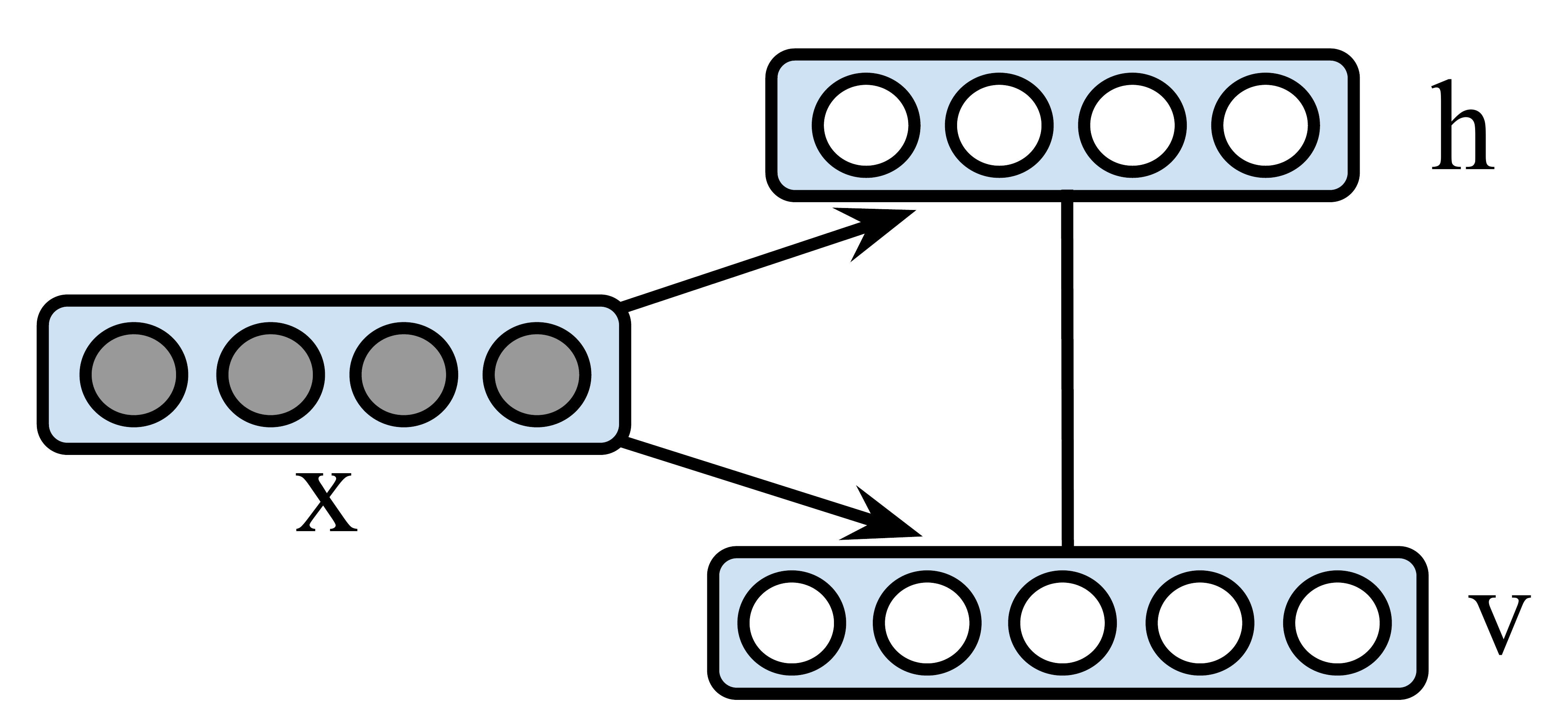}} \\
{\small (a) RBM} &  {\small (b) conditional RBM}
\end{tabular}
\vspace{-0.7em}
\caption{Graphical illustration of (a)~RBM with $|\vv v| = 5$ visible units and $|\vv h| = 4$ hidden units, 
	and 	(b)~the extended CRBM with $\vv v$ as output variables,  $\vv x$ as observed input features.
}
\label{fig:crbm}
\end{figure}

An RBM is a undirected graphical model (see Figure \ref{fig:crbm}(a)) 
that defines a joint distribution over the vectors of  
visible units $\vv v \in \{0, 1\}^{|\vv v| \times 1}$ and hidden units $\vv h \in \{ 0, 1\}^{|\vv h|\times 1}$, 
\begin{align}
\label{rbm}
p(\vv v, \vv h | \theta) = \frac{1}{Z(\theta)} 
\exp \big(  -E(\vv v, \vv h; \theta)   \big),
\end{align}
where $|\vv v|$ and $|\vv h|$ are the dimensions of $\vv v$ and $\vv h$ respectively;
$E(\vv v, \vv h; \theta)$ is the energy function, 
$$
 E(\vv v, \vv h; \theta) = -\vv v^\top W^{vh} \vv h  -  \vv v^\top \vv b^1 - \vv h^\top \vv b^2;
$$
 and $\theta = \{  W^{vh},  \vv b^1, \vv b^2 \}$ are the model parameters,
including pairwise interaction terms $W^{vh} \in \mathbb{R}^{|\vv v| \times |\vv h|}$, and
   bias terms $\vv b^1 \in \mathbb{R}^{|\vv v| \times 1}$ for visible units 
   and $\vv b^2 \in \mathbb{R}^{|\vv h| \times 1}$ for hidden units.
The function $Z(\theta)$ is the normalization constant, or \emph{partition function},
\begin{align*}
Z(\theta) = \sum_{\vv v} \sum_{\vv h} \exp \big(  -E(\vv v, \vv h; \theta)   \big),
\end{align*}
which is typically intractable to calculate. 


\subsection{Conditional RBM}
The conditional RBM~(CRBM) extends RBMs to include observed features $\vv x$
(see Figure \ref{fig:crbm}(b) 
for an illustration~\citep{mnih11}), 
and defines a joint conditional distribution over 
$\vv v$ and $\vv h$
given features $\vv x \in \mathbb{R}^{|\vv x| \times 1}$,
\begin{align}
\label{crbm}
p( \vv v, \vv h | \vv x ; \theta ) = \frac{1} {Z(\vv x; \theta)} \exp \big( - E( \vv v, \vv h, \vv x; \theta) \big),
\end{align}
where the energy function $E$ is defined as,
\begin{align*}
E( \vv v, \vv h, \vv x; \theta) &=   - {\vv v}^\top W^{vh} {\vv h} -  {\vv v}^\top W^{vx} {\vv x}  -  \vv h^\top W^{hx}  \vv x  \\
		&\qquad\qquad		- \vv v^{\top} \vv b^v - \vv h^\top \vv b^h, 
\end{align*}
and $\theta = \{  W^{vh}, W^{vx}, W^{hx},  b^v, b^h \}$ are model parameters. 
$Z(\vv x; \theta)$ is the $\vv x$-dependent partition function,
\begin{align*}
Z(\vv x; \theta) = \sum_{\vv v} \sum_{\vv h}  \exp \big( - E( \vv v, \vv h, \vv x; \theta) \big).
\end{align*} 
One can view an RBM as a special CRBM with $\vv x \equiv 0$.
\vspace{-1.5em}
\paragraph{Conditional Distribution}
Because CRBMs have a bipartite structure given the observed features, 
the conditional distributions $p(\vv v | \vv h, \vv x)$  and $p(\vv h | \vv v, \vv x)$ are fully 
factored and can be written as,
\begin{align}
\label{crbm_conditional}
\!\!\!\!
& p(\vv v | \vv h, \vv x) = \prod_{i} p( v_i |  \vv h, \vv x  ),  \ \ \
p(\vv h | \vv v, \vv x) = \prod_{j} p( h_j |  \vv v, \vv x  )	   \nonumber \\
& \text{with }
p( v_i =1 |  \vv h, \vv x   ) 
 = \sigma \big(  
  W^{vh}_{i \bullet} \vv h  
+  W^{vx}_{i \bullet} \vv x   + b^v_i \big),  \nonumber \\
& \qquad \
 p( h_j =1 |  \vv v, \vv x  ) 
 = \sigma \big( \vv v^{T} W^{vh}_{\bullet j}   +  W^{hx}_{j \bullet} \vv x   + b^h_j  \big) , 
\end{align}
where $\sigma (u) = 1 / (1 + \exp(-u))$ is the logistic function,
$W^{vh}_{i \bullet} $ and $ W^{vh}_{\bullet j} $  are the $i$-th row and $j$-th column of $W^{vh}$ respectively,
$W^{vx}_{i \bullet}$ is the $i$-th row of of  $W^{vx}$,
and $W^{hx}_{j \bullet }$ is the $j$-th row of  $W^{hx}$.
Eq.~\eqref{crbm_conditional} allows us to derive a blocked Gibbs sampler that iteratively alternates between drawing $\vv v$ and $\vv h$.

\vspace{-0.2em}
\paragraph{Marginal Distribution}
The marginal distribution of visible units $\vv v$ given observed features $\vv x$ is,
\begin{myequation}
\!\!\!\!\!
p( \vv v | \vv x ) = \sum_{\vv h} p(\vv v, \vv h | \vv x ) 
= \frac{1}{Z(\vv x; \theta)} \exp \big[ -F (\vv v, \vv x; \theta) \big]  
\!\!\!\!
\label{crbm_marginal}
\end{myequation}
where the negative 
energy function has analytic form,
\vspace{-1.5em}
{\small
\begin{align*}
-F (\vv v, \vv x; \theta) =  
& \sum_{j=1}^{|\vv h|}  \log \big[ 1 + \exp \big( \vv v^\top W_{\bullet j}^{vh}  + W_{j\bullet}^{hx} \vv x + b_j^h \big)  \big] \\
& \qquad + \vv v^\top W^{vx} \vv x  +  \vv v^\top \vv b^v .
\end{align*}
}
\!\!\!\!\!\!\!
 Note, after marginalizing out hidden variables, the log-linear model~\eqref{crbm} becomes a non-linear model~\eqref{crbm_marginal}, 
which can capture high-order correlations among visible units. 
This property is essentially important in many applications of CRBMs with structured output~\citep[e.g.,][]{salakhutdinov07, mnih11}.
\subsection{Structured Prediction with CRBMs}
In structured prediction, the visible units $\vv v$ typically represent output variables, 
while the observed $\vv x$ represent input features,
and the hidden units $\vv h$ facilitate the modeling of output variables given observed features.
To make predictions, one choice is to infer the modes of the singleton marginals,
$p(v_i | \vv x) =  \sum_{\vv v_{\backslash i}} \sum_{\vv h}  p(\vv v, \vv h | \vv x ).$
This marginalization inference is intractable and is closely related to calculating the partition function.
One can also decode the output $\vv v$ by performing joint maximum \emph{a posteriori}~(MAP) inference \citep[e.g.,][]{yu2009learning, yang14},
\begin{align*}
( \hat{\vv v},  \hat{ \vv h} )  = \argmax_{\vv v, \vv h}~  p(\vv v, \vv h | \vv x),
\end{align*}
which gives a prediction for the pair $(\vv v, \vv h)$; one obtains a prediction of $\vv v$ by simply discarding the $\vv h$ component.
Intuitively, the joint MAP prediction is ``over-confident'', since it deterministically assigns 
the hidden units to their most likely states,
and is not robust when the uncertainty of the hidden units is high.
One promising alternative for CRBM is marginal MAP prediction~\citep{ping14},
\begin{align*}
\tilde{\vv v}  = \argmax_{\vv v} p(\vv v | \vv x)  = \argmax_{\vv v} \sum_{\vv h} \exp\big(-E( \vv v, \vv h, \vv x; \theta) \big),
\end{align*}
which explicitly takes into account the uncertainty of the hidden units by marginalizing them out.
In general, these predictions are intractable in CRBMs, and one must use approximate inference methods, such as mean field or belief propagation.

\vspace{-0.1em}
\section{LEARNING}
\vspace{-0.3em}
\label{sec:learning}
In this section, we discuss different learning methods for conditional RBMs.
\vspace{-0.2em}
\subsection{MLE and Related Algorithms}
\vspace{-0.3em}
Assume we have a training set $\{\vv v^n, \vv x^n \}_{n=1}^N$; then, the log-likelihood can be written as,%
\begin{align*}
\sum_{n=1}^{N}  \Big\{ 
\log \sum_{\vv h} \exp \big( - E( \vv v^n, \vv h, \vv x^n; \theta) \big)  
 - \log Z(\vv x^n; \theta)	 
\Big\}.
\end{align*}
To efficiently maximize the objective function, stochastic gradient descent~(SGD) is usually applied.
Given a randomly chosen instance $\{ \vv v^n, \vv x^n \}$, 
one can show that the gradient of log-likelihood w.r.t.\ $W^{vh}$ is,
\begin{align}
\frac{\partial \log p(\vv v^{n} | \vv x^{n} ) } {\partial W^{vh} } 
= \vv v^{n} {(\vv \mu^n )}^\top  - \E_{ p( \vv v, \vv h | \vv x^{n} ) } \big[ \vv v \vv h^\top \big] ,
\label{loglike_grdt_Wvh}
\end{align}
where $\vv \mu^n = \sigma ( {W^{vh}}^\top  \vv v^n + W^{hx} \vv x^n + \vv b^h  )$
and the logistic function $\sigma$ is applied in an element-wise manner.
The positive part of the gradient can be calculated exactly.
The negative part arises from the derivatives of the log-partition function and is intractable to calculate.
The gradients of log-likelihood w.r.t.\ other parameters are analogous to Eq.~\eqref{loglike_grdt_Wvh}, and can be found in Appendix~A. 

CD-k initializes the Gibbs chain by instance $\vv v^{n}$, and performs $k$-step Gibbs sampling by Eq.~\eqref{crbm_conditional}.
Then, the empirical moment is used as a substitute for the intractable expectation 
$\E_{ p( \vv v, \vv h | \vv x^{n} ) } \big[ \vv v \vv h^\top \big]$.
Although this works well on RBMs, it gives unsatisfactory results on CRBMs.
In practice, the conditional distributions $p(\vv v, \vv h | \vv x^n)$ are strongly influenced by the observed features $\vv x^{n}$,
and usually more peaked than generative RBMs.
It is usually difficult for a Markov chain with few steps~(e.g., 10) to explore these peaked and multi-modal distributions.
PCD uses a long-run persistent Markov chain to improve convergence, but is not suitable for CRBMs as discussed in Section~\ref{sec:introduction}.

Sum-product BP and mean field methods provide pseudo-marginals as substitutes for the intractable expectations in Eq.~\eqref{loglike_grdt_Wvh}. 
These deterministic gradient estimates have the advantage that a larger learning rate can be used.
%
%
BP tends to give a more accurate estimate of $\log Z$ and marginals, but is reported to be slow on CRBMs  
and is impractical on problems with large output dimension and hidden layer sizes in structured prediction~\citep{mnih11}.

More importantly, it was observed that belief propagation usually gives unsatisfactory results when learning vanilla RBMs.
This is mainly because the parameters' magnitude gradually increases during learning; the RBM model eventually undergoes a ``phase transition'' after which BP has difficulty converging~\citep{ihler2005loopy,mooij2005properties}. 
If BP does not converge, it can not provide a meaningful gradient direction to update the model, and the leaning becomes stuck.
However, CRBMs appear to behave quite differently, due to operating in the ``high signal'' regime provided by
an informative observation $\vv x$.
This improves the convergence behaviour of BP, which may not be surprising since loopy BP is widely accepted as useful in learning other conditional models (e.g., grid CRFs for image segmentation).
In addition, given $N$ training instances for learning the CRBM,  BP is actually performed on $N$ different RBMs corresponding to different features $\vv x^n$.  During any particular phase of learning, 
BP may have trouble converging on some training instances, 
but we can still make progress as long as BP converges on the majority of instances.
%
We demonstrate this behavior in our experiments.

\vspace{-0.2em}
\subsection{Max-Margin Learning}
\vspace{-0.3em}
Another by-product of using BP is that it enables us to apply the  marginal structured SVM~(MSSVM)~\citep{ping14}
framework for max-margin learning of CRBMs,
\vspace{-1em}
\begin{align}
\label{mssvm}
\min_{\theta}\  \sum_{n=1}^{N}  &\Big\{  
\max_{\vv v} \log \sum_{\vv h} \exp \Big( \Delta( {\vv v, \vv v^{n}} )  - E( \vv v, \vv h, \vv x; \theta) \Big)  \nonumber \\
&- \log \sum_{\vv h} \exp \Big( - E( \vv v^n, \vv h, \vv x^n; \theta) \Big)  
 \Big\},
\end{align}
 \vspace{-1.5em}
\\ where the loss function $ \Delta ( \vv v, \vv v^{n} ) = \sum_i \Delta (v_i, v_i^{n})  $ is decomposable (e.g., Hamming loss).
In contrast to LSSVM \citep{yu2009learning, yang14},  MSSVM marginalizes over 
the uncertainty of hidden variables, and can significantly outperform LSSVM when that uncertainty is large~\citep{ping14}.
Experimentally, we find that 
MSSVM improves performance of max-margin CRBMs, likely because there is usually non-trivial uncertainty in the hidden units.
Given an instance $\{ \vv v^n, \vv x^n \}$, the stochastic gradient of Eq.~\eqref{mssvm}  w.r.t. $W^{vh}$ is,
\begin{align}
\label{mssvm_grdt}
\frac{\partial l(\vv v^{n} , \vv x^{n} ) } {\partial W^{vh} } 
=  \E_{ p( \vv h | \hat{\vv v}, \vv x^{n} ) } \big[ \hat{\vv v} \vv h^T \big]
	-  \vv v^{n} {(\vv \mu^n )}^\top,
\end{align}
where $\vv \mu^n$ is defined as in Eq.~\eqref{loglike_grdt_Wvh};
$\hat{\vv v}$ is the loss-augmented marginal MAP prediction,
$$
\hat{\vv v} = \argmax_{\vv v} \sum_{\vv h} \exp \Big( \Delta( {\vv v, \vv v^{(n)}} )  - E( \vv v, \vv h, \vv x^n; \theta) \Big);
$$ 
and ``mixed-product'' belief propagation~\citep{liu13} 
or dual-decomposition method~\citep{ping2015decomposition} 
for marginal MAP can provide 
pseudo-marginals to estimate the intractable expectation. 
(The gradients for other parameters are analogous.)

\section{APPROXIMATE INFERENCE}
\label{sec:inference}
\vspace{-0.1em}
In this section, we present a matrix-based implementation of sum-product and mixed-product BP algorithms for RBMs.
Given a particular $\vv x^n$ in CRBM~\eqref{crbm}, we obtain a $\vv x^n$-dependent RBM model, 
\vspace{-0.2em}
\begin{align*}
p(\vv v, \vv h | \vv x^n) = \frac{1}{Z(\theta(\vv x^n))} 
\exp \big( \vv v^\top W^{vh} \vv h  +  \vv v^\top \vv b^1 + \vv h^\top \vv b^2  \big),
\end{align*}
where the bias terms $\vv b^1 = \vv b^v + W^{vx} \vv x^n$,\ $\vv b^2 = \vv b^h + W^{hx} \vv x^n$, 
and thus we can directly apply the algorithm to CRBMs.
\vspace{-0.1em}
\subsection{Message-passing in RBMs}
\vspace{-0.1em}
We first review the standard message-passing form in RBMs.
On a dense graphical models like RBMs, to reduce the amount of calculation, 
one should always pre-compute the product of incoming messages~(or the beliefs) on the nodes, 
and reuse them to perform updates of all outgoing messages. 
In sum-product BP, we write the fixed-point update rule for the message sent from hidden unit $h_j$ to visible unit $v_i$ as,
\begin{align}
\label{m_hj_to_vi}
m_{j \rightarrow i} (v_i)  \propto \sum_{h_j}
\exp \big( v_i W^{vh}_{ij} h_j  \big)  \cdot \frac{ \tau (h_j) } { m_{i \rightarrow j} (h_j) } ,
\end{align}
where the belief on $h_j$ is
\begin{align}
\label{tau_hj}
\tau ( h_j ) & \propto \exp \big( h_j b^2_j \big) \cdot  \prod_{k=1}^{|\vv v|}  m_{k \rightarrow j} (h_j).
\end{align}
The update rule for the message sent from $v_i$ to $h_j$ is,
\begin{align}
\label{m_vi_to_hj}
m_{i \rightarrow j} (h_j ) \propto \sum_{v_i}
\exp \big( v_i W^{vh}_{ij} h_j  \big)  \cdot \frac{\tau (v_i)} { m_{j \rightarrow i} (v_i) },
\end{align}
where the belief on $v_i$ is,
\begin{align}
\label{tau_vi}
\tau ( v_i ) & \propto \exp \big( v_i b^1_i \big)  \cdot  \prod_{k=1}^{|\vv h|}  m_{k \rightarrow i} (v_i).
\end{align}
In mixed-product BP, the message sent from hidden unit to visible unit is the same as Eq.~\eqref{m_hj_to_vi}.
The message sent from visible unit $v_i$ to hidden unit $h_j$ is 
\begin{align}
\label{arg_max_m_vi_to_hj}
\tilde{m}_{i \rightarrow j} (h_j ) \propto 
\exp \big( \tilde{v_i} W^{vh}_{ij} h_j  \big)  \cdot \frac{\tau (\tilde{v_i})} { m_{j \rightarrow i} (\tilde{v_i}) },
\end{align}
where $\tilde{v_i} = \argmax_{v_i} \tau(v_i)$, and $\tau(v_i)$ is defined in Eq.~\eqref{tau_vi}.
These update equations are repeatedly applied until the values converge~(hopefully), or a stopping criterion is satisfied.
Then, the pairwise belief on $(v_i, h_j )$ is calculated as,
\begin{align*}
\tau (v_i, h_j ) & \propto  \exp\big( v_i W^{vh}_{ij} h_j \big)  \cdot
\frac{  \tau( v_i )  } { m_{j \rightarrow i} (v_i) }   \cdot  \frac{ \tau (h_j)  }{m_{i \rightarrow j} (h_j )} .
\end{align*}
%
%
It is well known that BP on loopy graphs is not guaranteed to converge, although in practice it usually does~\citep{murphy99loopyBP}.
%
%
\vspace{-0.1em}
\subsection{Matrix-based BP Algorithms}
\vspace{-0.1em}
Our algorithms use a compact matrix representation.
We denote the ``free'' belief vectors and matrices as,
\begin{align*}
&\vv \tau^v \in \R^{|\vv v| \times 1},  
	\text{ where }  \tau^v_i = \tau (v_i = 1),
\\
&\vv \tau^h \in \R^{|\vv h| \times 1}, 
	\text{ where }  \tau^h_j = \tau (h_j = 1), 
\\
&\Gamma \in \R^{ |\vv v| \times |\vv h| }, 
	\text{ where }   \Gamma_{ij}  = \tau(v_i=1, h_j=1). 
\end{align*}
Other beliefs can be represented by these ``free'' beliefs:
\begin{align*}
&\tau (v_i = 0) = 1 - \tau^v_i ,\quad  \tau (h_j = 0) = 1 - \tau^h_j,
\\
&\tau(v_i=1, h_j=0) = \tau^v_i - \Gamma_{ij}, 
\\
&\tau(v_i=0, h_j=1) = \tau^h_j - \Gamma_{ij}, 
\\
&\tau(v_i=0, h_j=0) = 1 + \Gamma_{ij} - \tau^v_i - \tau^h_j .
\end{align*}
We similarly define the normalized message matrices,
\begin{align*}
& M^{vh} \in \R^{|\vv v|\times |\vv h|} ,
 \quad  M^{vh}_{ij}  =  m_{j \rightarrow i} (v_i = 1) , 
\\
& M^{hv} \in \R^{|\vv h|\times |\vv v|} ,
 \quad  M^{hv}_{ji}  =  m_{i \rightarrow j} (h_j = 1) .
\end{align*}
Thus, $M^{vh}$ represents all the messages sent from $\vv h$ to $\vv v$,
and $M^{hv}$ represents all the messages from $\vv v$ to $\vv h$.

One can show (see Appendix B.1) 
that the update equation for message matrix $M^{vh}$ in both sum-product and mixed-product BP is
\begin{align}
\label{M_vh_update} 
M^{vh} =~ &\sigma \Big(
		\log \Big( \frac{\exp(W^{vh}) \circ \Lambda^{vh}_1 + \Lambda^{vh}_2 }  { \Lambda^{vh}_1 + \Lambda^{vh}_2 } 
		\Big)  \Big),
\\ \text{ where } \quad
& \Lambda^{vh}_1 = (\vv 1^{hv} - M^{hv})^\top \cdot \diag(\vv \tau^h) , \nonumber \\
& \Lambda^{vh}_2 = {M^{hv}}^\top  \cdot \diag(\vv 1^{h} - \vv \tau^h) , \nonumber
\end{align}
where $\vv 1^{hv}$ is a $|\vv h| \times |\vv v|$ matrix of ones, 
$\vv 1^h$ is a $|\vv h| \times 1$ vector of ones,
$\circ$ is the element-wise Hadamard product, 
and $\diag(\cdot)$ extracts the elements in a vector to form a diagonal matrix.
The logarithm,  fraction and logistic function are all applied in an element-wise manner.
Similarly, the update equation for message matrix $M^{hv}$ in sum-product BP is
\begin{align}
\label{M_hv_update} 
M^{hv} =~ &\sigma \Big(
		\log \Big( \frac{\exp({W^{vh}}^\top) \circ \Lambda^{hv}_1 + \Lambda^{hv}_2 }  { \Lambda^{hv}_1 + \Lambda^{hv}_2 }
		\Big)  \Big),
\\ \text{ where } \quad
& \Lambda^{hv}_1 = (\vv 1^{vh} - M^{vh})^\top \cdot \diag(\vv \tau^v) , \nonumber \\
& \Lambda^{hv}_2 = {M^{vh}}^\top  \cdot \diag(\vv 1^{v} - \vv \tau^v),	\nonumber
\end{align}
with $\vv 1^{vh}$ a $|\vv v| \times |\vv h|$ matrix of ones, 
and $\vv 1^v$  a $|\vv v| \times 1$ vector of ones.
In mixed-product BP, the update equation for message matrix $M^{vh}$ is
\begin{align}
\label{M_hv_update_mixedBP}
 {M}^{hv} = \sigma \Big(   {W^{vh}}^\top  \cdot  \diag(\tilde{\vv v} ) \Big),  
\end{align}
where  $\tilde{v_i} = \argmax_{v_i}  \tau^{v} (v_i)$ for all $v_i$.
In addition, one can show~(see Appendix B.2) 
that the belief vectors $\vv \tau^v$ and $\vv \tau^h$ can be calculated as,
\vspace{-0.3em}
\begin{align}
\label{tau_v_update}
\vv \tau^v  &= \sigma  \Big( \vv b^1 + \log \Big(  \frac{M^{vh}} {\vv 1^{vh}  - M^{vh}} \Big)  \cdot \vv 1^{h} \Big) , 
\\
\label{tau_h_update}
\vv \tau^h  &=  \sigma \Big( \vv b^2 + \log \Big(  \frac{M^{hv}} {\vv 1^{hv} - M^{hv}}  \big) \cdot \vv 1^{v} \Big) ,
\end{align}
\vspace{-0.3em}
\!\!where $\cdot$ is the matrix product.
These update equations are repeatedly applied until the stopping criterion is satisfied.
After that, the pairwise belief matrix $\Gamma$ can be calculated as,
\vspace{-0.3em}
\begin{align}
\!\!\!\!\!\!\!\!
\label{pairwise_belief}
\Gamma \ =  \frac{\Gamma^{11}} { \Gamma^{11} + \Gamma^{01} + \Gamma^{10} + \Gamma^{00}  },  \text{  where }
\qquad \qquad 
\end{align}
\vspace{-2em}
{ \small
\begin{align}
\Gamma^{11} &= \exp(W^{vh}) \circ (\vv \tau^v \cdot  {\vv \tau^h}^\top) \circ (\vv 1^{vh}-M^{vh}) \circ (\vv 1^{hv}-M^{hv})^\top ,  \nonumber
\\ %
\Gamma^{01} &= \big( (\vv 1^v - \vv \tau^v) \cdot {\vv \tau^h}^\top \big) \circ M^{vh} \circ (\vv 1^{hv} - M^{hv})^\top ,  \nonumber
\\ %
\Gamma^{10} &= \big(\vv \tau^v \cdot (\vv 1^h -\vv \tau^h)^\top \big) \circ (\vv 1^{vh} -M^{vh}) \circ {M^{hv}}^\top ,  \nonumber
\\ %
\Gamma^{00} &= \big( (\vv 1^v - \vv \tau^v) \cdot (\vv 1^h - \vv \tau^h) \big) \circ M^{vh} \circ {M^{hv}}^\top .  \nonumber
\end{align}
}
%
%
%
%
\!\!\!We summarize the matrix-based sum-product BP and mixed-product BP  in Algorithm~\ref{sumBP}.
It is well known that asynchronous~(sequential) BP message updates usually converge much faster than synchronous updates~\citep[e.g.,][]{wainwright2003tree, gonzalez2009residual}; 
in Algorithm~\ref{sumBP}, although messages are sent in parallel from all hidden units to visible units,  
the bipartite graph structure ensures that these are actually \emph{asynchronous} updates, which helps
convergence in practice.
%
Our method is also related to 
message-passing algorithms designed for other binary networks, such as binary LDPC codes
\citep{kschischang2001factor}, 
which parametrize each message by a single real number using a hyperbolic tangent transform. 
Our algorithm is specially designed for RBM-based models, and significantly speeds up BP by taking advantage of the RBM structure and using only matrix operations. 
In practice, our matrix implementation runs orders of magnitude faster than standard implementation of belief propagation, e.g., the C++ factor graph package libDAI~\citep{mooij2010libdai}, which has been used for RBM assessments~\citep[e.g.,][]{hadjis2015importance}.
For an RBM with 1000 visible and 500 hidden units, 10 iterations of BP in our Matlab implementation takes 
0.5 seconds on a laptop with Intel Core i5 (2.5GHz). In libDAI (with gcc -O3, i.e., fully optimized for speed), 10 iterations of BP takes 
297.4
seconds, approximately $600\times$ slower. 
This is mainly because matrix operations are highly optimized in modern computer architectures, e.g., they are performed in parallel in the instruction pipeline, and no pointers (to messages, neighbors, etc.) need to be dereferenced.

\begin{algorithm}[tb]
   \caption{Sum(mixed)-product BP on RBM}
   \label{sumBP}
\begin{algorithmic}
   \STATE {\bfseries Input:} 
   		$\{ W^{vh}, \vv b^1, \vv b^2 \}$, number of iterations $T$
   \STATE {\bfseries Output:}
     	beliefs $\{ \vv \tau^v, \vv \tau^h, \Gamma \}$
   \STATE initialize message matrices: \\
   				 \qquad $M^{vh} =0.5 \times \vv 1^{vh}$, \ \ $M^{hv} = 0.5\times\vv 1^{hv}$;
   \STATE initialize beliefs: 
   				 \ \  $\vv \tau^v = \sigma(\vv b^1)$, \ \ $\vv \tau^h = \sigma(\vv b^2)$;
   \FOR{$t=1$ {\bfseries to} $T$} 
  		\STATE send messages from $\vv h$ to $\vv v$:
   		  \STATE \quad $\Lambda^{vh}_1 = (\vv 1^{hv} - M^{hv})^\top \cdot \diag(\vv \tau^h)$;
		  \STATE \quad $\Lambda^{vh}_2 = {M^{hv}}^\top  \cdot \diag(\vv 1^{h} - \vv \tau^h)$;
		  \STATE \quad $ M^{vh} = \sigma \Big(
		 						\log \Big( \frac{\exp(W^{vh}) \circ \Lambda^{vh}_1 + \Lambda^{vh}_2 }  { \Lambda^{vh}_1 + \Lambda^{vh}_2 } 
								\Big)  \Big)$; \hfill \eqref{M_vh_update}
		  \STATE \quad $\vv \tau^v  = \sigma  \Big( \vv b^1 + \log \Big(  \frac{M^{vh}} {\vv 1^{vh}  - M^{vh}} \Big)  \cdot \vv 1^{h} \Big)$;
		  					 	\hfill \eqref{tau_v_update}
		\STATE send messages from $\vv v$ to $\vv h$:
			\STATE \quad  for sum-product BP
			\STATE \qquad\ $\Lambda^{hv}_1 = (\vv 1^{vh} - M^{vh})^\top \cdot \diag(\vv \tau^v)$;
			\STATE \qquad\ $\Lambda^{hv}_2 = {M^{vh}}^\top  \cdot \diag(\vv 1^{v} - \vv \tau^v)$;
		    \STATE \qquad\ $M^{hv} = \sigma \Big(
						\log \Big( \frac{\exp({W^{vh}}^\top) \circ \Lambda^{hv}_1 + \Lambda^{hv}_2 }  { \Lambda^{hv}_1 + \Lambda^{hv}_2 }
								\Big)  \Big)$; \hfill \eqref{M_hv_update}
			\STATE \quad  or, for mixed-product BP
			\STATE \qquad\ ${M}^{hv} = \sigma \Big(   {W^{vh}}^\top  \cdot  \diag( \tilde{\vv v} ) \Big)$; \hfill \eqref{M_hv_update_mixedBP}
		    \STATE \quad $\vv \tau^h = \sigma \Big( \vv b^2 + \log \Big(  \frac{M^{hv}} {\vv 1^{hv} - M^{hv}}  \big) \cdot \vv 1^{v} \Big)$; 
		    					\hfill  \eqref{tau_h_update}  \qquad\qquad
   \ENDFOR  \vspace{.5em}
   \STATE $\Gamma = \frac{\Gamma^{11}} { \Gamma^{11} + \Gamma^{01} + \Gamma^{10} + \Gamma^{00} }$\  
   						as defined in Eq.~\eqref{pairwise_belief};  
\end{algorithmic}
\end{algorithm}

\vspace{-0.5em}
\section{Experiments}
\vspace{-0.5em}
\label{sec:experiment}
In this section, we compare our methods with state-of-the-art algorithms for learning  CRBMs on two datasets:
MNIST and Caltech101 Silhouettes.

\begin{figure}[t] \centering
\vskip -0.1in
\centerline{\includegraphics[width=7.7 cm]{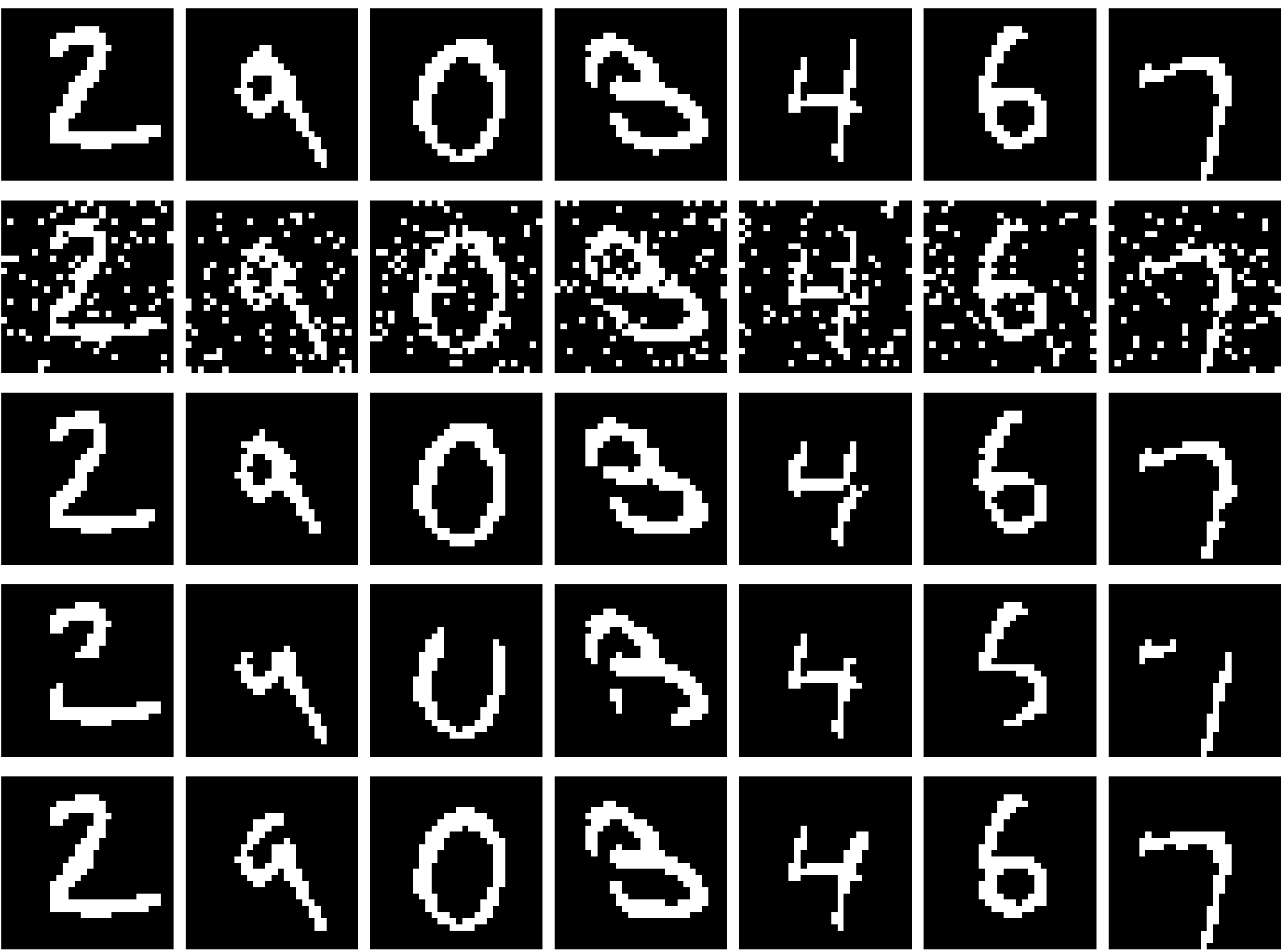}}
\vspace{-0.6em}
\caption{(Row 1) 7 original images from the test set. 
(Row 2)~The noisy~($10\%$) images. 
(Row 3)~The images predicted from noisy images. 
(Row 4)~The occluded~($8 \times 8$) images.
(Row 5)~The images predicted from the occluded images.
Rows 3 and 5 use our MLE-BP for learning.}
\vskip -0.14in
\label{fig:demo}
\end{figure}

{\bf Datasets:}
The MNIST 
database \citep{lecun1998gradient} contains $60,000$ images in the training set and $10,000$ test set images.
We randomly select $10,000$ images from training as the validation set.
Each image is $28 \times 28$ pixels, thus $|\vv v| = 784$.
We binarize the grayscale images by thresholding the pixels at 127, to obtain the clean image $\vv v$.
We test two types of structured prediction tasks in our experiment.
The first task is image denoising and denoted ``\emph{noisy} MNIST'', 
where the noisy image $\vv x$ is obtained by flipping either $10\%$ or $20\%$ of the entries in $\vv v$.
The second task is image completion, denoted \emph{occluded} MNIST, where the occluded image $\vv x$ is obtained by setting a random patch within the image $\vv v$ to 0.
The patch size is either $8 \times 8$ or $12\times 12$ pixels.
See Figure~\ref{fig:demo} for an illustration. \\
%
The Caltech101 Silhouettes dataset~\citep{marlin2010inductive} has $8,671$ images with $28 \times 28$ binary pixels, where each image represents object silhouette.
The dataset is divided into three subsets: $4,100$ examples for training, $2,264$ for validation and $2,307$ for testing. 
We test both image denoising and image completion tasks.
The noisy image $\vv x$ in \emph{noisy} Caltech101 is obtained by flipping $20\%$ of the pixels from the clean $\vv v$,
and the occluded image in \emph{occluded} Caltech101 is obtained by setting a random $12 \times 12$ patch to 1.

{\bf Model:}
Following \citep{mnih11}, 
we structured the CRBM model with $256$ hidden units,
giving $1$ million parameters in the model.
All the learning algorithms are applied to learn this CRBM model. 
The logistic regression method can be viewed as learning this CRBM with only $W^{vx}$ and $\vv b^{v}$ non-zero.

{\bf Algorithms:}
We train several CRBMs using the state-of-the-art CD methods, including CD-1, CD-10 and CD-PercLoss.
We also train models to optimize likelihood (MLE) using mean field~(MLE-MF) and sum-product BP~(MLE-BP).%
\footnote{In previous work~\citep{mnih11}, MLE-BP was considered impractical on this task 
due to the efficiency issue.}
Finally, we train MSSVM CRBMs using mixed-product BP, and LSSVM CRBMs using max-product BP.
A fixed learning rate is selected from the set $\{0.05, 0.02, 0.01, 0.005\}$ 
using the validation set,
and the mini-batch size is selected from the set $\{10, 20, 40, 80, 160\}$.
The CD-PercLoss algorithm uses $10$-step Gibbs sampling in the stochastic search process.
All the CD methods use 200 epochs in training.
In contrast, MLE-MF, MLE-BP, MSSVM and LSSVM use 50 epochs, because BP and MF provide a deterministic gradient estimate and larger learning rates can be applied.
Early stopping based on the validation error is also used for all methods.%
\footnote{In experiments, we found that early stopping always worked better than the Frobenius norm regularization.}
We test the learned models of the CD methods and MLE-MF with mean-field predictions;  the learned model of MLE-BP with sum-product BP predictions; MSSVM with mixed-product BP; and LSSVM with max-product BP.

\begin{table}[t]
\vskip -0.2in
\tabcolsep 1.5mm
\caption{Average test error~(\%) for image denoising on \emph{noisy} MINIST. 
All denotes the percentage incorrectly labeled pixels among all pixels.
Changed denotes the percentage of errors among pixels that were changed by the noise process.}
\begin{center}
    \begin{tabular}{c||cc|cc}
    \hline
    \qquad \ Dataset  &  \multicolumn{2}{c|}{Noisy~($10\%$)} &  \multicolumn{2}{c}{Noisy~($20\%$)}  \\ 
    Method \qquad \quad  & All 	& Changed & All & Changed  \\ \hline \hline
    LR & $1.960$  & $12.531$  & $4.088$  & $12.609$  \\ \hline
    CD-1 & $1.925$ & $12.229$ & $4.012$ &  $12.597$ \\ 
    CD-10 & $1.816$  &  $11.103$ &  $3.995$ & $11.271$ \\ 
    CD-PercLoss  & $1.760$ & $11.121$ & $3.970$ & $10.876$ \\
    MLE-MF  & $1.862$ & $11.319$ & $3.917$  &  $10.939$ \\ 
    MLE-BP  & ${\bf 1.688}$ & ${\bf 10.718}$ & ${\bf 3.691}$ & ${\bf 10.409}$ \\
    LSSVM  &	 $1.807$	& $11.565$  & $3.910$ &  $11.175$  \\ 
    MSSVM  &	 $1.751$	& $11.023$  & $3.804$  & $10.627$ \\ 	\hline
    \end{tabular}
\end{center}
\label{tab:corr_MNIST}
\vskip -0.05in
\tabcolsep 1.5mm
\caption{Average test error~(\%) for image completion on \emph{occluded} MINIST.}
\vspace{-0.7em}
\begin{center}
    \begin{tabular}{c||cc|cc}
    \hline
    \qquad \ Dataset  &  \multicolumn{2}{c|}{Occluded~($8\times8$)} &  \multicolumn{2}{c}{Occluded~($12\times12$)}  \\ 
    Method \qquad \quad  & All 	& Changed & All & Changed  \\ \hline \hline
    LR & $1.468$ & $61.304$  & $3.498$ & $53.971$ \\ \hline
    CD-1 & $1.814$ & $63.130$ &  $3.983$ & $58.376$\\ 
    CD-10 & $1.707$  &  $67.925$ & $3.921$ & $63.237$  \\ 
    CD-PercLoss & $1.394$ & $45.684$  & $3.483$ & $35.755$  \\
    MLE-MF  & $1.492$ & $49.553$ &  $3.477$ &  $40.703$  \\ 
    MLE-BP  & ${\bf 1.329}$  & ${\bf 39.785}$   &  ${\bf 3.117}$ & $36.233$ \\
    LSSVM  &	 $1.496$	& $44.037$  & $3.468$ &  $39.140$  \\ 
    MSSVM  &	 $1.391$	& $41.829$  & $3.273$  & ${\bf 35.712}$ \\ 	\hline
    \end{tabular}
\end{center}
\label{tab:occl_MNIST}
\vskip -0.15in
\end{table}
\begin{table}[t]
\vskip -0.2in
\tabcolsep 1.5mm
\caption{Average test error~(\%) for image denoising \& completion on Caltech101 Silhouettes dataset.}
\vspace{-0.8em}
\tabcolsep 1.5mm
\begin{center}
    \begin{tabular}{c||cc|cc}
    \hline
    \qquad \ Dataset  &  \multicolumn{2}{c|}{Noisy~($20\%$)} &  \multicolumn{2}{c}{Occluded~($12\times12$)}  \\ 
    Method \qquad \quad  & All 	& Changed & All & Changed  \\ \hline \hline
    LR & $5.653$ & $11.460$  & $4.771$ &  $16.587$ \\ \hline
    CD-1 & $5.876$ & $12.423$ &  $5.033$ & $20.300$\\ 
    CD-10 & $5.736$  &  $12.013$ & $5.149$ & $21.087$  \\ 
    CD-PercLoss & $5.622$ & $10.808$  & $5.081$ & $15.102$  \\
    MLE-MF  & $5.617$ & $11.083$ &  $4.692$ &  $15.995$  \\ 
    MLE-BP  & ${\bf 5.445}$  & ${\bf 10.731}$   &  $4.548$ & $16.541$ \\ 
    LSSVM  &	 $5.628$   & $11.468$  & $4.703$ &  $16.014$  \\ 
    MSSVM  & $5.549$   &   $11.389$  &  ${\bf 4.534}$  &  ${\bf 14.918}$  \\ \hline
    \end{tabular}
\end{center}
\label{tab:Caltech101}
\vskip -0.1in
\end{table}

{\bf Results:}
Table~\ref{tab:corr_MNIST} shows the percentage of incorrectly labeled pixels on the \emph{noisy} MNIST for different methods.
``All'' denotes the errors among all pixels and is the main measurement. 
We also report  the ``Changed'' errors among the pixels that were changed by the noise/occlusion process.
MLE-BP works best and provides $4\%$ and $7\%$ relative improvement over CD-PercLoss on two datasets with different noise levels.
Table~\ref{tab:occl_MNIST} shows the results on  \emph{occluded} MNIST.
Here MLE-BP provides $4\%$ and $10\%$ relative improvement over CD-PercLoss on the two datasets, respectively. 
CD-k gives unsatisfactory results in both cases.
Here MSSVM performs worse than MLE-BP, but better than the other methods in Table \ref{tab:corr_MNIST} and \ref{tab:occl_MNIST}.
The image completion task is viewed as more difficult on Changed pixels. However, again training the CRBM with MLE-BP gives very good results; see the last two rows of images in Figure~\ref{fig:demo}.  
Table~\ref{tab:Caltech101} demonstrate the results on Caltech101 Silhouettes; in this setting, 
MLE-BP and MSSVM perform the best for image denoising and image completion, respectively.

\begin{figure}[t] \centering
\vskip -0.08in
\includegraphics[width=5cm]{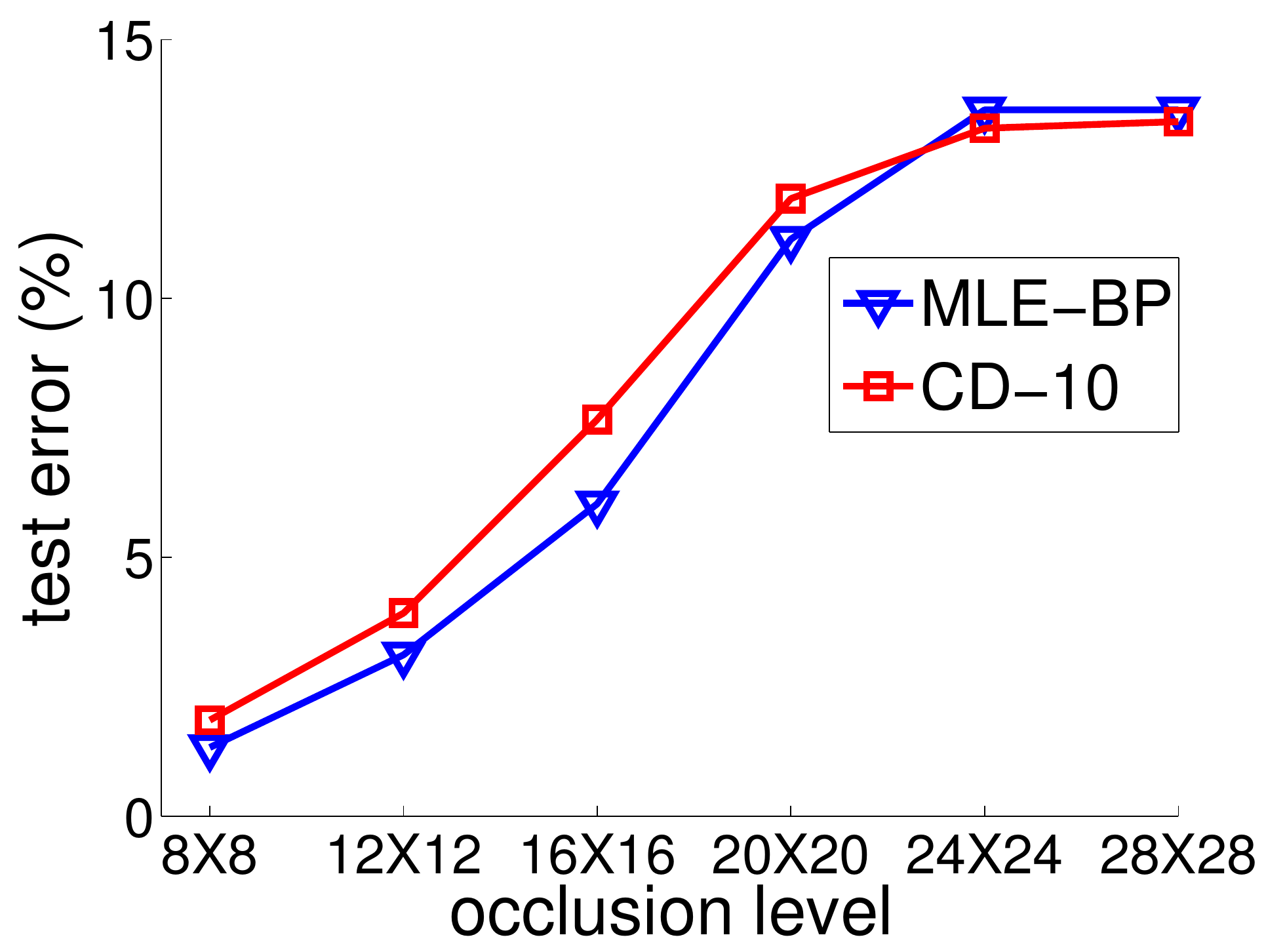}
\vspace{-1em}
\caption{ Average test error~(\%) for image completion on \emph{occluded} MINIST under different occlusion levels.
}
\vspace{-0.5em}
\label{fig:occl_level}
\end{figure}

Figure~\ref{fig:occl_level} shows the results for image completion under different occlusion levels. 
MLE-BP works better than CD-10, unless the images are almost fully occluded.
Note, the full occlusion~($28\times 28$) corresponds to no conditioning~(i.e., $\vv x \equiv 0$),
 and the CRBM models are reduce to vanilla RBMs.

\begin{figure}[t] \centering
\vskip -0.06in
\includegraphics[width=4.5cm]{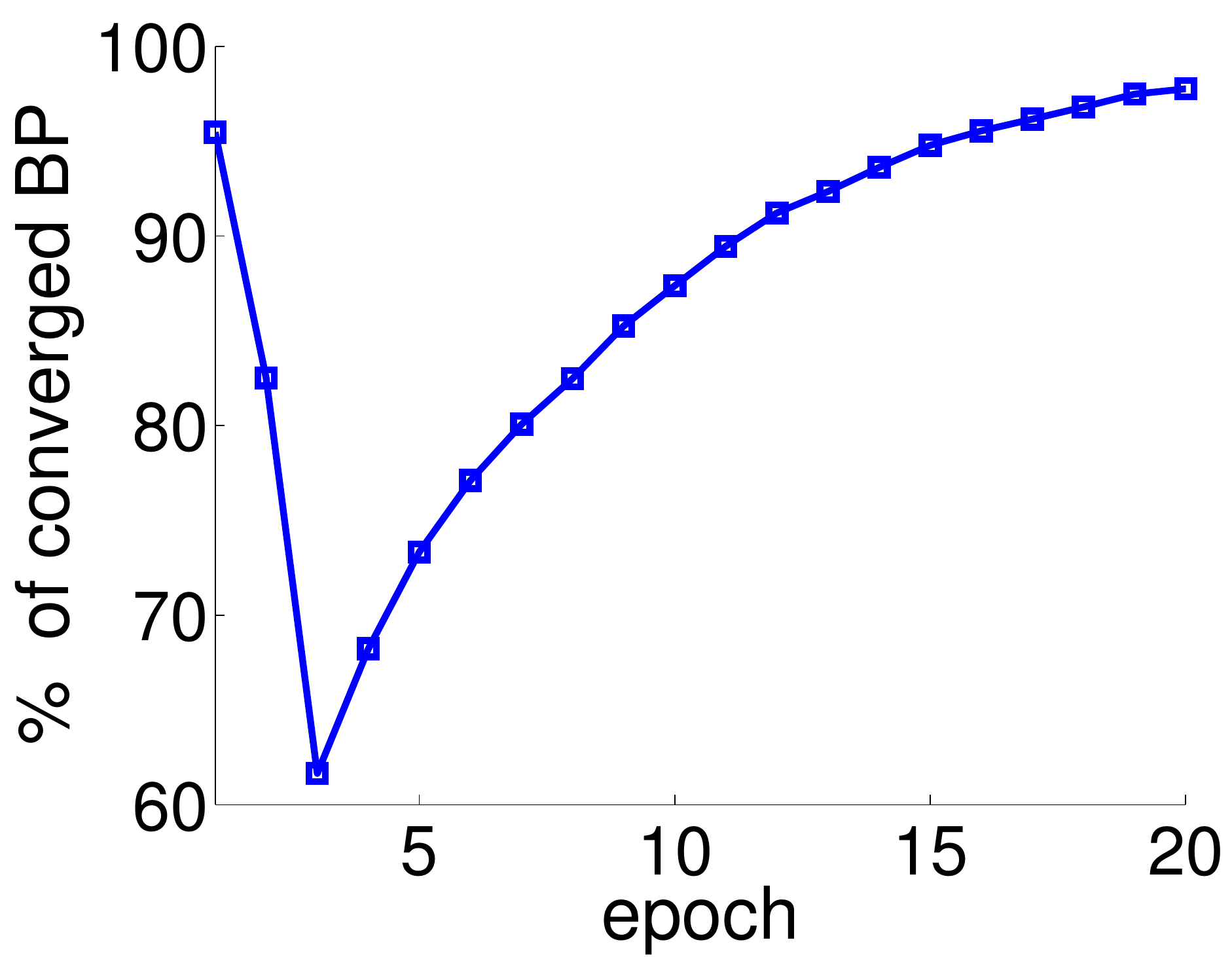}
\vspace{-1em}
\caption{ Percentage of converged BP in each epoch during MLE-BP training on \emph{occluded} ($8\times 8$) MNIST. 
}
\vspace{-0.5em}
\label{fig:conv_bp}
\end{figure}

{\bf Discussion:} 
We include several observations on the interaction of learning and inference algorithms for CRBMs:
(1)~Early on in learning, message passing is fast to converge, typically within $\approx 7$ iterations.
As learning continues, the magnitudes of the parameters gradually increase, and 
it becomes harder for BP to converge quickly.
One simple but effective strategy is to set the number of iterations to $T = 7 + \text{epoch}$~(e.g., at epoch 10,~$T = 17$).
See Figure~\ref{fig:conv_bp} for an illustration of the convergence behavior of BP using this strategy during training.
We set the convergence tolerance $\epsilon = 0.001$.
The model undergoes a change of convergence behaviour around epoch 3, but we can still make progress as BP converges on the majority of training instances.
(2)~No damping is better. Although message damping can improve the convergence of BP, it always requires more iterations of message-passing and effectively slows down the progress of learning CRBMs.
(3)~The approximate inference algorithms used in learning and test should be matched, 
which means the inference method~(BP or mean-field), number of iterations etc., should be the same.
Otherwise, we see unsatisfactory results.  
(4)~Learning CRBMs and vanilla RBMs are quite different in practice.  
As the literature suggests, in vanilla RBMs we also find that CD methods work better than MLE-BP.
%

%

%

\vspace{-0.5em}
\section{Conclusions and Future Work}
\label{sec:conclusion}
\vspace{-0.3em}
In contrast to past work, we argue that belief propagation can be an excellent 
choice for learning and inference with RBM-based models in the conditional setting.
We present a matrix-based expression of the BP updates for CRBMs,  
which is scalable to tens of thousands of visible and hidden units.
Our implementation takes advantage of the bipartite graphical structure and uses a 
compact representation of messages and beliefs.
Since it uses only matrix product and element-wise operations, 
it is highly suited to GPU acceleration.
We demonstrate that learning CRBMs with sum-product BP (MLE) and mixed-product BP (MSSVM) can 
provide significantly better 
results than the state-of-the-art CD methods on structured prediction problems.
Future directions include a GPU-based implementation and applying the method to deep probabilistic models, such as deep Boltzmann machines.

\vspace{-0.5em}
\subsubsection*{Acknowledgements}
\vspace{-0.3em}
This work is sponsored in part by NSF grants IIS-1254071 and CCF-1331915. It is also funded
in part by the United States Air Force under Contract No. FA8750-14-C-0011 under the DARPA
PPAML program.


\bibliographystyle{abbrvnat}
\bibliography{crbm}

\appendix

\section*{\Large Appendix}

\section{Gradients of Log-likelihood}
\label{appendix_grdt}
Similar to Eq.~(5)  
in the main text, the gradients of log-likelihood with other weights and biases are,
\begin{align*}
\frac{\partial \log p(\vv v^{n} | \vv x^{n} ) } {\partial W^{vx} } 
&= \vv v^{n}  {\vv x^{n}}^\top
	- \E_{  p( \vv v, \vv h | \vv x^{n} ) } \big[ \vv v   {\vv x^{n}}^\top  \big],		\\
\frac{\partial \log p(\vv v^{n} | \vv x^{n} ) } {\partial W^{hx} } 
	&= \vv \mu^n   {\vv x^{n}}^\top    -  \E_{ p( \vv v, \vv h | \vv x^{n} ) } \big[ \vv h  {\vv x^{n}}^\top  \big], \\
\frac{\partial \log p(\vv v^{n} | \vv x^{n} ) } {\partial b^{v} } 
&= \vv v^{n} 
	- \E_{ p( \vv v, \vv h | \vv x^{n} ) } \big[ \vv v  \big],		\\
\frac{\partial \log p(\vv v^{n} | \vv x^{n} ) } {\partial b^{h} } 
&= \vv\mu^n
	- \E_{ p( \vv v, \vv h | \vv x^{n} ) } \big[ \vv h  \big],
\end{align*}
where $\vv \mu^n = \E_{ p( \vv h | \vv v^n, \vv x^n ) } \big[ \vv h \big]  =  \sigma ( {W^{vh}}^\top \vv v^n + W^{hx} \vv x^n + \vv b^h  )$.
All the negative parts of these gradients are intractable to calculate, and must be approximated during learning.

\section{Derivation of Matrix-based BP}
\label{appendix_update}
In this section we give additional proof details of our matrix-based BP update equations.

{\bf B.1}~ 
Proof of the update rule for $M^{vh}$ in Eq.~(13):  
{\small
\begin{align*}
M^{vh}_{ij} &= \frac{ m_{j \rightarrow i} (v_i = 1) } {m_{j \rightarrow i} (v_i = 1) + m_{j \rightarrow i} (v_i = 0)  } , 
	\\
&= \sigma \Big(  \log \frac { \exp(W^{vh}_{ij} ) \cdot \frac{ \tau^h_j } { M^{hv}_{ji} }  + \frac{1-\tau^h_j}{1-M^{hv}_{ji}}  }  
									{  \frac{ \tau^h_j } { M^{hv}_{ji} }  + \frac{1-\tau^h_j}{1-M^{hv}_{ji}}    }  \Big) , 
									\text{ by Eq.~(8) } 
    \\ 
&= \sigma \Big( \log \frac{ \exp(W^{vh}_{ij} ) \cdot (1 - M^{hv}_{ji}) \tau^h_j + M^{hv}_{ji} (1-\tau^h_j)   }
									{  (1 - M^{hv}_{ji}) \tau^h_j + M^{hv}_{ji} (1-\tau^h_j)   } \Big) .
\end{align*}
}
\!\!\!\!\! Then, one can verify  the update of $M^{vh}$~(13)  
holds. 
The derivation is analogous for updating $M^{hv}$~(14).  

%
%
\noindent
{\bf B.2}~Proof of the update rule for $\vv \tau^v$ in  Eq.~(16): 
{\small
\begin{align*}
\tau^v_i &= \frac{\tau (v_i = 1)} {\tau (v_i = 1) + \tau (v_i = 0)} , 
		\\
	&= \frac{ 1 }  {
					 1 +  \frac { \exp \big( 0 + \sum_{j=1}^{|\vv h|} \log m_{j\rightarrow i}(v_i=0)  \big) }   
					 { \exp \big(b_i  +  \sum_{j=1}^{|\vv h|} \log m_{j\rightarrow i}(v_i=1) \big) } 
			} ,	   
			\text{  by Eq.~(11) }  
		\\
	&=  \frac{1} { 1+ \exp\big\{ - b^1_i - \sum_{j=1}^{|\vv h|} \big(  \log M^{vh}_{ij} - \log (1-M^{vh}_{ij}) \big) \big\}  } , 
		\\
	&=  \sigma \Big(   b^1_i + \big(  \log M^{vh}_{i\bullet} - \log (\vv {1^h}^\top - M^{vh}_{i\bullet}) \big) \cdot \vv 1^h \big\} \Big) .
\end{align*}
}
\!\!\! Then, one can verify the update of $\vv \tau^v$~(16) 
holds.
The update of $\vv \tau^h$ in Eq.~(17) 
is derived similarly. \vspace{0.3em}

%
\noindent
{\bf B.3}~The $(i, j)$ element of pairwise belief matrix:
{\small
\begin{align*}
&\Gamma_{ij} = \frac{\tau(v_i=1, h_j=1)}{ \sum_{v_i, h_j} \tau(v_i, h_j) } =   
\\
& \frac{ \frac{  \exp(w^{vh}_{ij}) \tau^v_i \tau^h_j } { M^{vh}_{ij} M^{hv}_{ji} }}
 	 {  \frac{ \exp(w^{vh}_{ij}) \tau^v_i \tau^h_j} { M^{vh}_{ij} M^{hv}_{ji} } 
 	 		+  \frac{ (1-\tau^v_i)\tau^h_j } { (1 - M^{vh}_{ij}) M^{hv}_{ji} }
 			+  \frac{ \tau^v_i (1 - \tau^h_j) } { M^{vh}_{ij} (1 - M^{hv}_{ji}) }
 			+  \frac{ (1-\tau^v_i) (1-\tau^h_j) } { (1-M^{vh}_{ij}) (1-M^{hv}_{ji}) }	
	 }	 
\end{align*} 
}
\!\!\!\!\! We can denote the intermediate terms
{ \small
\begin{align}
\Gamma^{11} &= \exp(W^{vh}) \circ (\vv \tau^v \cdot  {\vv \tau^h}^\top) \circ (\vv 1^{vh}-M^{vh}) \circ (\vv 1^{hv}-M^{hv})^\top ,  \nonumber
\\ %
\Gamma^{01} &= \big( (\vv 1^v - \vv \tau^v) \cdot {\vv \tau^h}^\top \big) \circ M^{vh} \circ (\vv 1^{hv} - M^{hv})^\top ,  \nonumber
\\ %
\Gamma^{10} &= \big(\vv \tau^v \cdot (\vv 1^h -\vv \tau^h)^\top \big) \circ (\vv 1^{vh} -M^{vh}) \circ {M^{hv}}^\top ,  \nonumber
\\ %
\Gamma^{00} &= \big( (\vv 1^v - \vv \tau^v) \cdot (\vv 1^h - \vv \tau^h) \big) \circ M^{vh} \circ {M^{hv}}^\top .  \nonumber
\end{align}
}
\!\!\!\!\! Then, the pairwise belief matrix
$
\Gamma = \frac{\Gamma^{11}} { \Gamma^{11} + \Gamma^{01} + \Gamma^{10} + \Gamma^{00}  }.
$

\end{document}